\documentclass[10pt,twocolumn,letterpaper]{article}

\usepackage{iccv}
\usepackage{times}
\usepackage{epsfig}
\usepackage{graphicx}
\usepackage{amsmath}
\usepackage{amssymb}

\usepackage{graphicx}
\graphicspath{ {./images/} }
\usepackage[pagebackref=true,breaklinks=true,letterpaper=true,colorlinks,bookmarks=false]{hyperref}

\iccvfinalcopy % *** Uncomment this line for the final submission

\def\iccvPaperID{46} % *** Enter the ICCV Paper ID here

\ificcvfinal\fi

\begin{document}

%%%%%%%%% TITLE
\title{Comics for Everyone: Generating Accessible Text Descriptions for Comic Strips}

\author{Reshma Ramaprasad\\
Microsoft Research\\
{\tt\small reshma.ramaprasad@microsoft.com}
% \and
}

\maketitle
\ificcvfinal\thispagestyle{empty}\fi

%%%%%%%%% ABSTRACT
\begin{abstract}
   Comic strips are a popular and expressive form of visual storytelling that can convey humor, emotion, and information. However, they are inaccessible to the BLV (Blind or Low Vision) community, who cannot perceive the images, layouts, and text of comics. Our goal in this paper is to create natural language descriptions of comic strips that are accessible to the visually impaired community. Our method consists of two steps: first, we use computer vision techniques to extract information about the panels, characters, and text of the comic images; second, we use this information as additional context to prompt a multimodal large language model (MLLM) to produce the descriptions. We test our method on a collection of comics that have been annotated by human experts and measure its performance using both quantitative and qualitative metrics. The outcomes of our experiments are encouraging and promising.
\end{abstract}

%%%%%%%%% BODY TEXT
\section{Introduction}

%-------------------------------------------------------------------------
\subsection{Accessible comics}

Comic strips are a form of visual storytelling that use humor, satire, and irony to convey a message or a joke. They usually have a series of panels or frames that show the actions and dialogues of the characters, setting of the scene etc. Comic strips are a popular medium that can attract a wide range of audiences. However, they are inaccessible to the BLV community, who cannot perceive the images, layouts, and text of comics. To make them accessible, we generate text descriptions that capture their humor and meaning. These descriptions can then be accessed via screen reader software for the visually impaired community. Comic strips vary widely in their style and content, and we need to select a representative sample for our experiments and evaluation. Therefore, we choose three popular comic strips that have different themes, characters, and humor: Dilbert, Garfield, and Peanuts. There are various elements of comic strips that need to be described to provide an authentic comic reading such as the count of panels, layout of panels, character, dialogue, context or setting, expressions, actions of the characters etc. While there is no standardized method for describing comic strips, ~\cite{cbd}  proposed a set of guidelines for comic book descriptions which we adopt for comic strips as well. 

\subsection{Multimodal large language model (MLLM)}

GPT4~\cite{gpt4} has sparked a research frenzy by highlighting the impressive capabilities of the model in not only processing text, but also the ability to understand and explain images, such as being able to generate a fully functioning website from a hand drawn picture of its design or explaining why a picture is funny. While the vision capabilities of GPT4 are still not accessible to the public, there have been various open- source efforts to develop capable models~\cite{survey}. Multimodal large language models (MLLMs) are neural networks that have displayed impressive capabilities in processing both natural language and visual information. In other words, they facilitate having conversations about images. We conduct our experiments on the LLaVA model~\cite{llava}, which is a state-of-the-art MLLM, but our findings can be generalized to other similar models. We believe that an MLLM based approach is suitable for this task as it is a conversational interface that can not only generate a description but further allow users to interact with the comic strips, such as asking questions about their humor or meaning.  

To the best of our knowledge, this is the first work that explores using MLLMs for the task of generating accessible text descriptions for newspaper comic strips. This is a challenging task because there are no clear rules for describing comic strips, there are few annotated datasets available for training on this domain, and comic strips have a lot of variety in their layouts, styles, and fonts, which make image processing difficult. Our main contributions are: 
\begin{enumerate}
    \item We propose a novel application of MLLMs for generating comic descriptions that can benefit people with visual impairments. 
    \item We identify the key elements of comic strips that are essential for generating good descriptions and propose a training-free, prompt-based approach that can improve the performance of MLLMs in this task. 
    \item  We evaluate our approach and provide insights into the strengths and weaknesses of our method. 
\end{enumerate}

%------------------------------------------------------------------------
\section{Related work }

%-------------------------------------------------------------------------
\subsection{Accessible comics}
 The design of an accessible comic book reader for the visually impaired community is the main objective of the studies conducted by ~\cite{access_comics,access_comics_2,alcove}. They explored the needs and preferences of this community through various user studies, and proposed different design aspects for the reader. An interactive web toon reader for the BLV community was also designed by ~\cite{cocomix} to enhance their reading experience. ~\cite{cbd} provided guidelines to help standardise the way comic book descriptions are written.
 Researchers ~\cite{index,mtl,cnn} have proposed various deep learning methods for extracting individual elements in comics like panels, characters, balloons etc,
Our work is the first, to the best of our knowledge, that leverages MLLMs to generate natural language descriptions for comics in an end-to-end manner.

%-------------------------------------------------------------------------
\subsection{Multimodal large language model (MLLM)}

The applications of MLLMs are wide from performing multimodal tasks such as VQA, image captioning, visual reasoning, writing stories from images, etc. They provide a friendly conversational interface which allows one to have engaging conversations with images. While these models have shown great potential in understanding and explaining images, they still struggle with text-related visual tasks~\cite{ocr}, such as reading the text in comic strips. This is a crucial requirement in our task of describing comics, as the dialogue of the characters is essential for an authentic comic reading experience. 
%-------------------------------------------------------------------------

\section{Proposed approach}
\begin{figure*}
\begin{center}
\includegraphics[width=15cm, height=5cm]{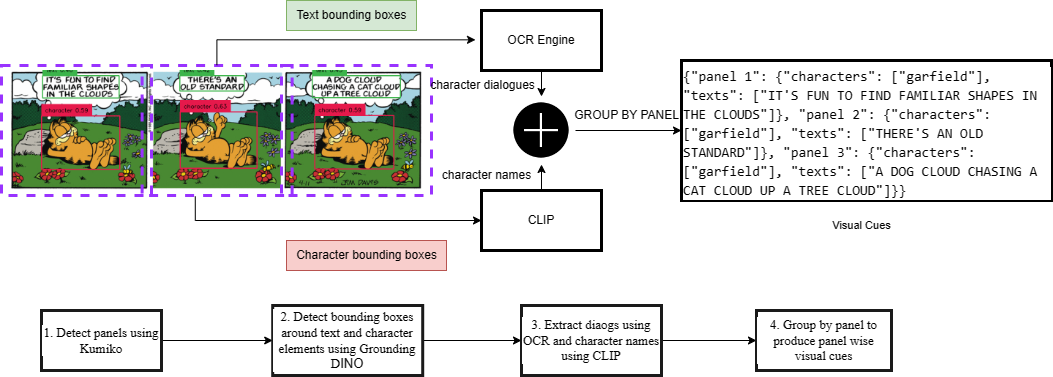}
\end{center}
   \caption{Visual cues extraction pipeline}
\label{fig:pipeline}
\end{figure*}
Figure \ref{fig:pipeline} illustrates the steps of the pipeline for extracting the visual cues (panel, text and character information) from a comic strip image. These visual cues are then used to create the context that we will pass as part of the prompt to the MLLM. .
\subsection{ Panel extraction}
To analyze the comic strips, we first need to extract the panels that contain the visual and textual information. For this purpose we use an open source tool called Kumiko~\cite{kumiko}. Kumiko uses the contour detection method provided by the openCV library to identify and separate the panels from the comic strip image. This method works well on comics where the panels have a white or black background separating them, which is common in many popular comic strips. We tested Kumiko on a collection of 60 single row comics from Dilbert, Garfield and Peanuts, and found that it performed well in extracting the panels accurately. 

\subsection{Character and text detection}
We use Grounding DINO~\cite{groundingdino}, a state-of-the-art open-set object detector, to detect the text and character elements in comics. Grounding DINO can detect arbitrary objects with human inputs such as category names or referring expressions. 
We pass the labels "text" and "character" to Grounding DINO as text prompts, which guide the model to generate bounding boxes around the corresponding visual elements in the comics. To control the quality of the object detection, Grounding DINO uses two parameters: TEXT\_THRESHOLD and BOX\_THRESHOLD. TEXT\_THRESHOLD determines how confident the model is that the text description matches the detected object. BOX\_THRESHOLD decides how precise the bounding box is around the object. We choose 0.2 as the value for both parameters, which means we accept moderate confidence and precision levels.  
We refine the bounding boxes from Grounding DINO by filtering out large text boxes (>80\% image area) as they are likely to be false positives. Second, we handle overlapping bounding boxes by using non-maximum suppression (NMS), a technique that removes redundant boxes based on their intersection over union (IoU) ratio. We use an IoU threshold of 0.5 for NMS. 
We believe that Grounding DINO can be a useful tool for automatic annotation of comics and training a specialized model for character and text detection. In this work, however, we use the output of Grounding DINO model directly without further fine-tuning.

\subsection{Character identification}
For character identification, we use CLIP~\cite{clip}, a multimodal large language model that can perform zero-shot image classification based on natural language prompts. CLIP is trained on a large number of pairs of images and their captions, and it can learn to associate visual concepts with natural language descriptions. This enables CLIP to recognize any object or category given its name, without requiring any labeled examples.  CLIP can also generalize to unseen characters or categories, as long as they have a meaningful name or description. 

We pass each character image detected in the previous step to CLIP for character identification. To use CLIP for character identification, we need to provide it with some prompts or labels for each character that we want to identify. These prompts are hand-crafted by us, based on the physical description of the characters. For example, we use “a boy with a yellow shirt and a round head” for Charlie Brown from Peanuts, or “a fat orange cat” for Garfield. 

However, there could be some collisions or overlaps of the labels for similar characters in different comics. For example, both Snoopy from Peanuts and Dogbert from Dilbert are white dogs, and adding more details to their labels, such as “a white dog with black ears” or “a small white dog who wears glasses”, might not be enough to distinguish them. To handle this case, we separate out the labels for each comic rather than pass all labels as categories to CLIP. This way, we can improve the accuracy of the zero-shot classification by reducing the ambiguity.

We believe that CLIP can be useful for automatically annotating comic characters and training a smaller specialized model. In this work, however, we use the output of CLIP model directly without further fine-tuning, as we want to preserve the generality of the model.

\subsection{Text identification (OCR)}
We use optical character recognition (OCR) to extract the text dialogue of the characters from the text bounding boxes that we detected in step 2. OCR is a technique that converts images of text into editable and searchable text. However, OCR quality can be affected by the stylized fonts used in comic strips, which may not be recognized by the OCR engine. We use Azure ACS OCR, which is a cloud-based service that provides high-quality OCR results on various types of images. We achieved good results on OCR with Azure ACS OCR on the comic strips that we experimented with.

%-------------------------------------------------------------------------

\section{Results}
In this section we present the results of character and text detection using Grounding DINO, character classification using CLIP as well as examples of the final comic description output. 

Our dataset consists of 60 comic strips from three popular comic series: Dilbert, Garfield, and Peanuts. We randomly sample 20 comic strips from each series that were published between 2022 and 2023.

\subsection{Character and text detection}
We applied the Grounding DINO model to our dataset of 60 comic strips and manually corrected the annotations to obtain the ground truth. The ground truth consists of 333 character bounding boxes and 238 text bounding boxes. We evaluated the accuracy of the bounding boxes detected by the model using the mean average precision (mAP) metric, which measures how well the model identifies the objects in the images. The mAP score was calculated using a threshold of 0.5 for the Intersection over Union (IoU) metric. The model achieved high scores for both character and text detection, with 92.14\% and 95.22\% average precision respectively. The overall mAP score was 93.68\%, indicating a good performance of the model.
%-------------------------------------------------------------------------

\subsection{Character identification}
We create the ground truth dataset by cropping the character images from the 333 bounding boxes that Grounding DINO detected. We use CLIP to automatically label the character images, and then we manually fix any errors. We used three metrics to measure the performance of our model: F1 score, precision, and recall. These metrics are computed using the functions f1\_score, precision\_score, and recall\_score from the sklearn library. The average=‘weighted’ parameter is passed to these functions to account for the class imbalance in the data. This means that each metric is calculated as a weighted average of the metrics for each class, where the weights are proportional to the number of samples in each class. The model achieved an F1 score of 0.7917, a precision of 0.8241, and a recall of 0.8258 on the test set.
%-------------------------------------------------------------------------

\subsection{Prompting MLLM with enhanced prompt}
\begin{figure*}[h!]
\begin{center}
\includegraphics[width=7.5cm, height=2.5cm]{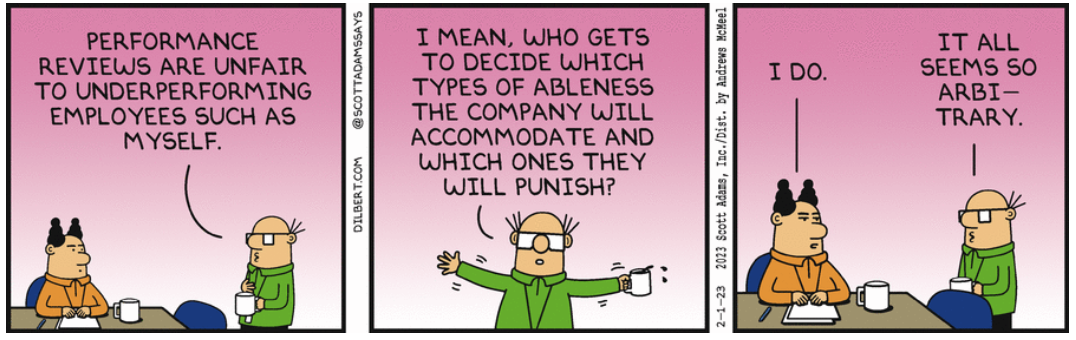}
\end{center}
   \caption{Sample Dilbert comic used as input to the LLaVA model.}
\label{fig:dilbert}
\end{figure*}
We use LLaVA-v1-13B-336px variant of the LLaVA model as the MLLM for our experiments. LLaVA~\cite{llava} is a large language and vision assistant that can generate text based on both words and images. It is trained by fine-tuning LLaMA/Vicuna on GPT-generated multimodal instruction-following data. LLaVA has different variants based on the size of the model, the resolution of the image, and the version of Vicuna that it is fine-tuned on. LLaVA-v1-13B-336px is a variant that has 13 billion parameters, can process images with 336x336 pixels, and is fine-tuned on Vicuna-v1. 

We use two prompts: a base prompt (P1) and an enhanced prompt (P2). P1 has only the comic image and a text asking the model to describe each panel. P2 also has the visual cues from the comic image that we extracted. We explain P1 and P2 below.

\textbf{P1 (base prompt)}: ``\emph{Your task is to write a text description for each panel of a comic strip. A comic strip is a sequence of drawings that tell a story using humor, satire, or irony. Each panel shows a scene with the characters, objects, actions, and dialogues. Your description should include the following elements:The number of panels in the comic strip.
The names and appearances of the characters in each panel.
The objects and background details in each panel.
The actions and expressions of the characters in each panel.
The dialogues or captions of the characters in each panel, using quotation marks or parentheses.
}"

The enhanced prompt is constructed using the base prompt as follows:

\textbf{P2 (enhanced prompt)}: \emph{P1 + ``Use the information given below to describe the characters and their dialogues and use your own knowledge to describe the other elements" + \{context\}}

We present an example output using the Dilbert comic shown in Figure \ref{fig:dilbert} first with the base prompt and then with our enhanced and show that the latter works better on this example. \\

\noindent \textbf{Case 1:} Using base prompt \\
\noindent \textbf{Prompt}: P1 \\
\textbf{Output:} \emph{The comic strip consists of three panels.
Panel 1:
In the first panel, there are two men sitting at a table. One man is holding a cup of coffee, while the other man is holding a cup of tea. The man with the cup of coffee is saying, "I mean, I'm not saying I'm perfect, but I'm not a complete failure either." The man with the cup of tea responds, "I'm not sure which is worse, being a failure or being a success."
Panel 2:
In the second panel, the man with the cup of coffee is saying, "I'm not sure which is worse, being a failure or being a success." The man with the cup of tea responds, "I'm not sure which is worse, being a failure or being a success." The two men are sitting at a table, and there is a clock on the wall.
Panel 3:
In the third panel, the man with the cup of coffee is saying, "I mean, I'm not saying I'm perfect, but I'm not a complete failure either." The man with the cup of tea responds, "I'm not sure which is worse, being a failure or being a success." The two men are sitting at a table, and there is a clock on the wall.
In summary, the comic strip consists of three panels, each featuring two men sitting at a table and discussing their thoughts on success and failure. The men are holding cups of coffee and tea, and there is a clock on the wall in each panel. The dialogues in the panels are humorous and satirical, as the men express their confusion and uncertainty about the concept of success and failure.} \\

\noindent \textbf{Case 2:} Using context-enhanced prompt \\
\textbf{Prompt:} P1 + Use the information given below to describe the characters and their dialogues and use your own knowledge to describe the other elements.
 \{"panel 1": {"characters": 
 ["wally", "the boss"],
 "texts": 
 ["PERFORMANCE REVIEWS ARE UNFAIR TO
 UNDERPERFORMING EMPLOYEES SUCH AS MYSELF."]},
 "panel 2": {"characters": ["wally"], 
 "texts": ["I MEAN, WHO GETS TO DECIDE 
 WHICH TYPES OF ABLENESS THE COMPANY WILL
 ACCOMMODATE AND WHICH ONES THEY WILL PUNISH?"]}, "panel 3": {"characters": 
 ["the boss", "wally"], 
 "texts": ["I DO", "IT ALL SEEMS SO ARBI- TRARY."]}\} \\
\textbf{Output:} \emph{The comic strip consists of three panels, each featuring a different scene and dialogue.
In the first panel, we see two characters, Wally and the Boss, standing in an office setting. Wally is holding a cup of coffee and expressing his frustration with the performance review system. He says, "PERFORMANCE REVIEWS ARE UNFAIR TO UNDERPERFORMING EMPLOYEES SUCH AS MYSELF." 
In the second panel, Wally is shown alone, still holding his cup of coffee, and expressing his confusion about the company's decision-making process. He asks, "I MEAN, WHO GETS TO DECIDE WHICH TYPES OF ABLENESS THE COMPANY WILL ACCOMMODATE AND WHICH ONES THEY WILL PUNISH?" 
In the third panel, Wally and the Boss are shown together, with the Boss responding to Wally's question. The Boss says, "I DO," and Wally adds, "IT ALL SEEMS SO ARBITRARY."
In summary, the comic strip features Wally, a frustrated employee, and the Boss, who is trying to explain the company's decision-making process. The dialogues highlight the challenges and complexities of performance reviews and the subjective nature of the company's decisions. }

Our approach shows promising results, but it also has failure cases where the comic descriptions are of low quality even with the enhanced prompt. We list some of the possible reasons for these failures:
\begin{enumerate}
  \item In our method we do not pass the character-dialogue association information as part of the context. Therefore, the model has to guess who is speaking what, which can lead to incorrect or inconsistent outputs. Sometimes the model also latches onto the ordering of characters and dialogue given to it in the context leading to incorrect outputs.
  \item The model sometimes makes up information that is not present or relevant in the comic strip image, such as adding extra panels, dialogues, or details. This can be attributed to the poor instruction-following capability and hallucination effects of the model.
  \item If the token limit of the model is too small then the context we pass in is ignored.
\end{enumerate}

%-------------------------------------------------------------------------

\section{Conclusion}
In this work we described a prompt-based method to produce natural language descriptions of comic strips using state of the art models. We used computer vision techniques and optical character recognition to extract information from the comic strip images, such as the panels, characters, and text. We then used this information as additional context to prompt a multimodal large language model (MLLM) to generate the descriptions. We evaluated our method on a dataset of 60 comic strips from three popular comic series: Dilbert, Garfield, and Peanuts. We compared our method with a baseline method that used only the comic strip image and a base prompt. We found that our method improved the quality and relevance of the descriptions. We believe our work establishes a generalisable and promising direction for automatically describing diverse comic strips. 
%-------------------------------------------------------------------------

\section{Acknowledgments}
The author expresses her gratitude to Saikat Guha and Tanuja Ganu for their valuable support and feedback.
%-------------------------------------------------------------------------

{\small
\bibliographystyle{ieee_fullname}
\bibliography{egbib}

\begin{thebibliography}{10}\itemsep=-1pt

\bibitem{kumiko}
{Kumiko the Comics Cutter}.
\newblock \url{https://github.com/njean42/kumiko}.

\bibitem{cnn}
David Dubray and Jochen Laubrock.
\newblock Deep cnn-based speech balloon detection and segmentation for comic books, 2019.

\bibitem{cocomix}
Mina Huh, YunJung Lee, Dasom Choi, Haesoo Kim, Uran Oh, and Juho Kim.
\newblock Cocomix: Utilizing comments to improve non-visual webtoon accessibility.
\newblock In {\em Proceedings of the 2022 CHI Conference on Human Factors in Computing Systems}, CHI '22, New York, NY, USA, 2022. Association for Computing Machinery.

\bibitem{access_comics}
Yunjung Lee, Hwayeon Joh, Suhyeon Yoo, and Uran Oh.
\newblock Accesscomics: An accessible digital comic book reader for people with visual impairments.
\newblock In {\em Proceedings of the 18th International Web for All Conference}, W4A '21, New York, NY, USA, 2021. Association for Computing Machinery.

\bibitem{access_comics_2}
Yun~Jung Lee, Hwayeon Joh, Suhyeon Yoo, and Uran Oh.
\newblock Accesscomics2: Understanding the user experience of an accessible comic book reader for blind people with textual sound effects.
\newblock {\em ACM Trans. Access. Comput.}, 16(1), mar 2023.

\bibitem{llava}
Haotian Liu, Chunyuan Li, Qingyang Wu, and Yong~Jae Lee.
\newblock Visual instruction tuning, 2023.

\bibitem{groundingdino}
Shilong Liu, Zhaoyang Zeng, Tianhe Ren, Feng Li, Hao Zhang, Jie Yang, Chunyuan Li, Jianwei Yang, Hang Su, Jun Zhu, and Lei Zhang.
\newblock Grounding dino: Marrying dino with grounded pre-training for open-set object detection, 2023.

\bibitem{ocr}
Yuliang Liu, Zhang Li, Hongliang Li, Wenwen Yu, Yang Liu, Biao Yang, Mingxin Huang, Dezhi Peng, Mingyu Liu, Mingrui Chen, Chunyuan Li, Xucheng Yin, Cheng lin Liu, Lianwen Jin, and Xiang Bai.
\newblock On the hidden mystery of ocr in large multimodal models, 2023.

\bibitem{index}
Nhu-Van Nguyen, Christophe Rigaud, and Jean-Christophe Burie.
\newblock Digital comics image indexing based on deep learning.
\newblock {\em Journal of Imaging}, 4(7):89, Jul 2018.

\bibitem{mtl}
Nhu-Van Nguyen, Christophe Rigaud, and Jean-Christophe Burie.
\newblock Comic mtl: optimized multi-task learning for comic book image analysis.
\newblock {\em International Journal on Document Analysis and Recognition (IJDAR)}, 22, 09 2019.

\bibitem{gpt4}
OpenAI.
\newblock Gpt-4 technical report, 2023.

\bibitem{cbd}
Rachel Osolen and Leah Brochu.
\newblock Creating an authentic experience: A study in comic books, accessibility, and the visually impaired reader.
\newblock {\em The International Journal of Information, Diversity and Inclusion}, 4(1):108--118, 2020.

\bibitem{clip}
Alec Radford, Jong~Wook Kim, Chris Hallacy, Aditya Ramesh, Gabriel Goh, Sandhini Agarwal, Girish Sastry, Amanda Askell, Pamela Mishkin, Jack Clark, Gretchen Krueger, and Ilya Sutskever.
\newblock Learning transferable visual models from natural language supervision, 2021.

\bibitem{alcove}
Fr\'{e}d\'{e}ric Rayar, Bernard Oriola, and Christophe Jouffrais.
\newblock Alcove: An accessible comic reader for people with low vision.
\newblock In {\em Proceedings of the 25th International Conference on Intelligent User Interfaces}, IUI '20, page 410–418, New York, NY, USA, 2020. Association for Computing Machinery.

\bibitem{survey}
Shukang Yin, Chaoyou Fu, Sirui Zhao, Ke Li, Xing Sun, Tong Xu, and Enhong Chen.
\newblock A survey on multimodal large language models, 2023.

\end{thebibliography}
}

\end{document}